# Mean Field Variational Approximation for Continuous-Time Bayesian Networks

**Ido Cohn**　　**Tal El-Hay**　　**Nir Friedman**
School of Computer Science
The Hebrew University
{ido_cohn,tale,nir}@cs.huji.ac.il

**Raz Kupferman**
Institute of Mathematics
The Hebrew University
raz@math.huji.ac.il

## Abstract

*Continuous-time Bayesian networks* is a natural structured representation language for multi-component stochastic processes that evolve continuously over time. Despite the compact representation, inference in such models is intractable even in relatively simple structured networks. Here we introduce a mean field variational approximation in which we use a product of *inhomogeneous* Markov processes to approximate a distribution over trajectories. This variational approach leads to a globally consistent distribution, which can be efficiently queried. Additionally, it provides a lower bound on the probability of observations, thus making it attractive for learning tasks. We provide the theoretical foundations for the approximation, an efficient implementation that exploits the wide range of highly optimized ordinary differential equations (ODE) solvers, experimentally explore characterizations of processes for which this approximation is suitable, and show applications to a large-scale real-world inference problem.

## 1 Introduction

Many real-life processes can be naturally thought of as evolving continuously in time. Examples cover a diverse range, including server availability, changes in socio-economic status, and genetic sequence evolution. To realistically model such processes, we need to reason about systems that are composed of multiple components (e.g., many servers in a server farm, multiple residues in a protein sequence) and evolve in continuous time. Continuous-time Bayesian networks (CTBNs) provide a representation language for such processes, which allows to naturally exploit sparse patterns of interactions to compactly represent the dynamics of such processes [9].

Inference in multi-component temporal models is a notoriously hard problem [1]. Similar to the situation in discrete time processes, inference is exponential in the number of components, even in a CTBN with sparse interactions [9]. Thus, we have to resort to approximate inference methods. The recent literature has adapted several strategies from discrete graphical models to CTBNs. These include sampling-based approaches, where Fan and Shelton [5] introduced a likelihood-weighted sampling scheme, and more recently we [4] introduced a Gibbs-sampling procedure. Such sampling-based approaches yield more accurate answers with the investment of additional computation. However, it is hard to bound the required time in advance, tune the stopping criteria, or estimate the error of the approximation. An alternative class of approximations is based on *variational principles*.

Recently, Nodelman et al. [11] introduced an *Expectation Propagation* approach, which can be roughly described as a local message passing scheme, where each message describes the dynamics of a single component over an interval. This message passing procedure can automatically refine the number of intervals according to the complexity of the underlying system [14]. Nonetheless, it does suffer from several caveats. On the formal level, the approximation has no convergence guaranties. Second, upon convergence, the computed marginals do not necessarily form a globally consistent distribution. Third, it is restricted to approximations in the form of piecewise-homogeneous messages on each interval. Thus, the refinement of the number of intervals depends on the fit of such homogeneous approximations to the target process. Finally, the approximation of Nodelman *et al* does not provide a provable approximation on the likelihood of the observation—a crucial component in learning procedures.

Here, we develop an alternative variational approximation, which provides a different trade-off. We use the strategy of structured variational approximations in graphical models [8], and specifically by the variational approach of Opper and Sanguinetti [12] for approximate inference in Markov Jump Processes, a related class of models (see below). The resulting procedure approximates the posterior distribution of the CTBN as a product of independent components, each of which is an inhomogeneous continuous-time Markov process. As we show, by using a natural representation of these processes, we derive a variational procedure that is both efficient, and provides a good approximation both for the likelihood of the evidence and for the expected sufficient statistics. In particular, the approximation provides a lower-bound on the likelihood, and thus is attractive for use in learning.

## 2 Continuous-Time Bayesian Networks

Consider a $D$-component Markov process $\boldsymbol{X}^{(t)} = (X_1^{(t)}, X_2^{(t)}, \ldots X_D^{(t)})$ with state space $S = S_1 \times S_2 \times \cdots \times S_D$. A notational convention: vectors are denoted by boldface symbols, e.g., $\boldsymbol{X}$, and matrices are denoted by blackboard style characters, e.g., $\mathbb{Q}$. The states in $S$ are denoted by vectors of indexes, $\boldsymbol{x} = (x_1, \ldots, x_D)$. We use indexes $1 \leq i, j \leq D$ for enumerating components and $\boldsymbol{X}^{(t)}$ and $X_i^{(t)}$ to denote the random variable describing the state of the process and its $i$'th components at time $t$.

The dynamics of a *time-homogeneous continuous-time Markov process* are fully determined by the *Markov transition function*,

$$p_{\boldsymbol{x},\boldsymbol{y}}(t) = \Pr(\boldsymbol{X}^{(t+s)} = \boldsymbol{y} | \boldsymbol{X}^{(s)} = \boldsymbol{x}),$$

where time-homogeneity implies that the right-hand side does not depend on $s$. These dynamics are fully captured by a matrix $\mathbb{Q}$—the *rate matrix* with non-negative off-diagonal entries $q_{\boldsymbol{x},\boldsymbol{y}}$ and diagonal $q_{\boldsymbol{x},\boldsymbol{x}} = -\sum_{\boldsymbol{y} \neq \boldsymbol{x}} q_{\boldsymbol{x},\boldsymbol{y}}$. This rate matrix defines the transition probabilities

$$p_{\boldsymbol{x},\boldsymbol{y}}(h) = \delta_{\boldsymbol{x},\boldsymbol{y}} + q_{\boldsymbol{x},\boldsymbol{y}} \cdot h + o(h)$$

where $\delta_{\boldsymbol{x},\boldsymbol{y}}$ is a multivariate Kronecker delta and $o(\cdot)$ means decay to zero faster than its argument. Using the rate matrix $\mathbb{Q}$, we can express the Markov transition function as $p_{\boldsymbol{x},\boldsymbol{y}}(t) = [\exp(t\mathbb{Q})]_{\boldsymbol{x},\boldsymbol{y}}$ where $\exp(t\mathbb{Q})$ is a matrix exponential [2, 7].

A *continuous-time Bayesian network* is defined by assigning each component $i$ a set of components $\mathbf{Pa}_i \subseteq \{1, \ldots, D\} \setminus \{i\}$, which are its parents in the network [9]. With each component $i$ we then associate a set of conditional rate matrix $\mathbb{Q}_{\cdot | \boldsymbol{u}_i}^{i | \mathbf{Pa}_i}$ for each state $\boldsymbol{u}_i$ of $\mathbf{Pa}_i$. The off-diagonal entries $q_{x_i, y_i | \boldsymbol{u}_i}^{i | \mathbf{Pa}_i}$ represent the rate at which $X_i$ transitions from state $x_i$ to state $y_i$ given that its parents are in state $\boldsymbol{u}_i$. The dynamics of $\boldsymbol{X}^{(t)}$ are defined by a rate matrix $\mathbb{Q}$ with entries $q_{\boldsymbol{x},\boldsymbol{y}}$, which amalgamates the conditional rate matrices as follows:

$$q_{\boldsymbol{x},\boldsymbol{y}} = \begin{cases} q_{x_i, y_i | \boldsymbol{u}_i}^{i | \mathbf{Pa}_i} & \delta(\boldsymbol{x}, \boldsymbol{y}) = \{i\} \\ \sum_i q_{x_i, x_i | \boldsymbol{u}_i}^{i | \mathbf{Pa}_i} & \boldsymbol{x} = \boldsymbol{y} \\ 0 & \text{otherwise,} \end{cases} \quad (1)$$

where $\delta(\boldsymbol{x}, \boldsymbol{y}) = \{i | x_i \neq y_i\}$. This definition implies that changes are one component at a time.

Given a continuous-time Bayesian network, we would like to evaluate the likelihood of evidence, to compute the probability of various events given the evidence (e.g., that the state of the system at time $t$ is $\boldsymbol{x}$), and to compute conditional expectations (e.g., the expected amount of time $X_i$ was in state $x_i$). Direct computations of these quantities involve matrix exponentials of the rate matrix $\mathbb{Q}$, whose size is exponential in the number of components, making this approach infeasible beyond a modest number of components. We therefore have to resort to approximations.

## 3 Variational Principle for Continuous Time Markov Processes

We start by defining a variational approximations principle in terms of a general continuous-time Markov process (that is, without assuming any network structure). For convenience we restrict our treatment to a time interval $[0, T]$ with end-point evidence $\boldsymbol{X}^{(0)} = \boldsymbol{e}_0$ and $\boldsymbol{X}^{(T)} = \boldsymbol{e}_T$. We discuss more general types of evidence below. Here we aim to define a lower bound on $\ln P_{\mathbb{Q}}(\boldsymbol{e}_T | \boldsymbol{e}_0)$ as well as to approximate the posterior probability $P_{\mathbb{Q}}(\cdot | \boldsymbol{e}_0, \boldsymbol{e}_T)$.

**Marginal Density Representation** Variational approximations cast inference as an optimization problem of a functional which approximates the log probability of the evidence by introducing an auxiliary set of *variational parameters*. Here we define the optimization problem over a set of *mean parameters* [15], representing possible values of expected sufficient statistics.

As discussed above, the prior distribution of the process can be characterized by a time-independent rate matrix $\mathbb{Q}$. It is easy to show that if the prior is a Markov process, then the posterior is also a Markov process, albeit not necessarily a homogeneous one. Such a process can be represented by a time-dependent rate matrix that describes the instantaneous transition rates. Here, rather than representing the target distribution by a time-dependent rate matrix, we consider a representation that is more natural for variational approximations. Let $\Pr$ be the distribution of a Markov process. We define a family of functions:

$$\mu_{\boldsymbol{x}}(t) = \Pr(\boldsymbol{X}^{(t)} = \boldsymbol{x})$$
$$\gamma_{\boldsymbol{x},\boldsymbol{y}}(t) = \lim_{h \downarrow 0} \frac{\Pr(\boldsymbol{X}^{(t)} = \boldsymbol{x}, \boldsymbol{X}^{(t+h)} = \boldsymbol{y})}{h}, \quad \boldsymbol{x} \neq \boldsymbol{y} \quad (2)$$
$$\gamma_{\boldsymbol{x},\boldsymbol{x}}(t) = -\sum_{\boldsymbol{y} \neq \boldsymbol{x}} \gamma_{\boldsymbol{x},\boldsymbol{y}}(t).$$

The function $\mu_{\boldsymbol{x}}(t)$ is the probability that $\boldsymbol{X}^{(t)} = \boldsymbol{x}$. The function $\gamma_{\boldsymbol{x},\boldsymbol{y}}(t)$ is the probability density that $\boldsymbol{X}$ transitions from state $\boldsymbol{x}$ to $\boldsymbol{y}$ at time $t$. Note that this parameter is not a transition rate, but rather a product of a pointwise probability with the point-wise transition rate of the approximating probability, i.e., $\gamma_{x,y}(t)/\mu_x(t)$ is the $\boldsymbol{x}, \boldsymbol{y}$ entry of the time-dependent rate matrix. Hence, unlike the (inhomogeneous) rate matrix at time $t$, $\gamma_{x,y}(t)$ takes into account the probability of being in state $\boldsymbol{x}$ and not only the rate of transitions. This definition implies that

$$\Pr(\boldsymbol{X}^{(t)} = \boldsymbol{x}, \boldsymbol{X}^{(t+h)} = \boldsymbol{y}) = \mu_{\boldsymbol{x}}(t)\delta_{\boldsymbol{x},\boldsymbol{y}} + \gamma_{\boldsymbol{x},\boldsymbol{y}}(t)h + o(h),$$

We aim to use the family of functions $\mu$ and $\gamma$ as a representation of a Markov process. To do so, we need to characterize the set of constraints that these functions should satisfy.

**Definition 3.1:** A family $\eta = \{\mu_{\boldsymbol{x}}(t), \gamma_{\boldsymbol{x},\boldsymbol{y}}(t) : 0 \leq t \leq T\}$ of continuous functions is a *Markov-consistent density set* if the following constraints are fulfilled:

$$\begin{aligned}
\mu_{\boldsymbol{x}}(t) &\geq 0, \quad \sum_{\boldsymbol{x}} \mu_{\boldsymbol{x}}(0) = 1, \\
\gamma_{\boldsymbol{x},\boldsymbol{y}}(t) &\geq 0 \quad \forall \boldsymbol{y} \neq \boldsymbol{x}, \\
\gamma_{\boldsymbol{x},\boldsymbol{x}}(t) &= -\sum_{\boldsymbol{y} \neq \boldsymbol{x}} \gamma_{\boldsymbol{x},\boldsymbol{y}}(t), \\
\frac{d}{dt} \mu_{\boldsymbol{x}}(t) &= \sum_{\boldsymbol{y}} \gamma_{\boldsymbol{y},\boldsymbol{x}}(t).
\end{aligned}$$

Let $\mathcal{M}$ be the set of all Markov-consistent densities. ∎

Using standard arguments we can show that there exists a correspondence between (generally inhomogeneous) Markov processes and density sets $\eta$. Specifically:

**Lemma 3.2:** *Let $\eta = \{\mu_{\boldsymbol{x}}(t), \gamma_{\boldsymbol{x},\boldsymbol{y}}(t)\}$. If $\eta \in \mathcal{M}$, then there exists a continuous-time Markov process $P_\eta$ for which $\mu_{\boldsymbol{x}}$ and $\gamma_{\boldsymbol{x},\boldsymbol{y}}$ satisfy (2).*

The processes we are interested in, however, have additional structure, as they correspond to the posterior distribution of a time-homogeneous process with end-point evidence. This additional structure implies that we should only consider a subset of $\mathcal{M}$:

**Lemma 3.3:** *Let $\mathbb{Q}$ be a rate matrix, and $\boldsymbol{e}_0, \boldsymbol{e}_T$ be states of $\boldsymbol{X}$. Then the representation $\eta$ corresponding to the posterior distribution $P_\mathbb{Q}(\cdot | \boldsymbol{e}_0, \boldsymbol{e}_T)$ is in the set $\mathcal{M}_{\boldsymbol{e}} \subset \mathcal{M}$ that contains Markov-consistent density sets satisfying $\mu_{\boldsymbol{x}}(0) = \delta_{\boldsymbol{x},\boldsymbol{e}_0}$, $\mu_{\boldsymbol{x}}(T) = \delta_{\boldsymbol{x},\boldsymbol{e}_T}$.*

Thus, from now on we can restrict our attention to density sets from $\mathcal{M}_{\boldsymbol{e}}$. The constraint that $\mu_{\boldsymbol{x}}(0)$ and $\mu_{\boldsymbol{x}}(T)$ also has consequences on $\gamma_{\boldsymbol{x},\boldsymbol{y}}$ at these points.

**Lemma 3.4:** *If $\eta \in \mathcal{M}_{\boldsymbol{e}}$ then $\gamma_{\boldsymbol{x},\boldsymbol{y}}(0) = 0$ for all $\boldsymbol{x} \neq \boldsymbol{e}_0$ and $\gamma_{\boldsymbol{x},\boldsymbol{y}}(T) = 0$ for all $\boldsymbol{y} \neq \boldsymbol{e}_T$.*

**Variational Principle**  We can now state the variational principle for continuous processes, which closely tracks similar principles for discrete processes.

We define a *free energy functional*,

$$\mathcal{F}(\eta; \mathbb{Q}) = \mathcal{E}(\eta; \mathbb{Q}) + \mathcal{H}(\eta),$$

which, as we will see, measures the quality of $\eta$ as an approximation of $P_\mathbb{Q}(\cdot|\boldsymbol{e})$. (For succinctness, we will assume that the evidence $\boldsymbol{e}$ is clear from the context.) The two terms in the continuous functional correspond to an entropy,

$$\mathcal{H}(\eta) = \int_0^T \sum_{\boldsymbol{x}} \sum_{\boldsymbol{y} \neq \boldsymbol{x}} \gamma_{\boldsymbol{x},\boldsymbol{y}}(t)[1 + \ln \mu_{\boldsymbol{x}}(t) - \ln \gamma_{\boldsymbol{x},\boldsymbol{y}}(t)] dt,$$

and an energy,

$$\mathcal{E}(\eta; \mathbb{Q}) = \int_0^T \sum_{\boldsymbol{x}} \left[ \mu_{\boldsymbol{x}}(t) q_{\boldsymbol{x},\boldsymbol{x}} + \sum_{\boldsymbol{y} \neq \boldsymbol{x}} \gamma_{\boldsymbol{x},\boldsymbol{y}}(t) \ln q_{\boldsymbol{x},\boldsymbol{y}} \right] dt.$$

**Theorem 3.5:** *Let $\mathbb{Q}$ be a rate matrix, $\boldsymbol{e} = (\boldsymbol{e}_0, \boldsymbol{e}_T)$ be states of $\boldsymbol{X}$, and $\eta \in \mathcal{M}_{\boldsymbol{e}}$. Then*

$$\mathcal{F}(\eta; \mathbb{Q}) = \ln P_\mathbb{Q}(\boldsymbol{e}_T|\boldsymbol{e}_0) - \boldsymbol{D}(P_\eta(\cdot) \| P_\mathbb{Q}(\cdot|\boldsymbol{e}))$$

*where $\boldsymbol{D}(P_\eta(\cdot) \| P_\mathbb{Q}(\cdot|\boldsymbol{e}))$ is the KL divergence between the two processes.*

We conclude that $\mathcal{F}(\eta; \mathbb{Q})$ is a lower bound of the log-likelihood of the evidence, and that the closer the approximation to the target posterior, the tighter the bound.

**Proof Outline**  The basic idea is to consider discrete approximations of the functional. Let $K$ be an integer. We define the $K$-sieve $\boldsymbol{X}_K$ to be the set of random variables $\boldsymbol{X}^{(t_0)}, \boldsymbol{X}^{(t_1)}, \ldots, \boldsymbol{X}^{(t_K)}$ where $t_k = \frac{kT}{K}$. We can use the variational principle [8] on the marginal distributions $P_\mathbb{Q}(\boldsymbol{X}_K|\boldsymbol{e})$ and $P_\eta(\boldsymbol{X}_K)$. More precisely, define

$$\mathcal{F}_K(\eta; \mathbb{Q}) = \mathbf{E}_{P_\eta} \left[ \ln \frac{P_\mathbb{Q}(\boldsymbol{X}_K, \boldsymbol{e}_T \mid \boldsymbol{e}_0)}{P_\eta(\boldsymbol{X}_K)} \right],$$

which can, by using simple arithmetic manipulations, be recast as

$$\mathcal{F}_K(\eta; \mathbb{Q}) = \ln P_\mathbb{Q}(\boldsymbol{e}_T|\boldsymbol{e}_0) - \boldsymbol{D}(P_\eta(\boldsymbol{X}_K) \| P_\mathbb{Q}(\boldsymbol{X}_K|\boldsymbol{e})).$$

We get the desired result by letting $K \to \infty$. By definition $\lim_{K \to \infty} \boldsymbol{D}(P_\eta(\boldsymbol{X}_K) \| P_\mathbb{Q}(\boldsymbol{X}_K|\boldsymbol{e}))$ is $\boldsymbol{D}(P_\eta(\cdot) \| P_\mathbb{Q}(\cdot|\boldsymbol{e}))$. The crux of the proof is in proving the following lemma.

**Lemma 3.6:** $\mathcal{F}(\eta; \mathbb{Q}) = \lim_{K \to \infty} \mathcal{F}_K(\eta; \mathbb{Q}).$

**Proof:** Since both $P_\mathbb{Q}$ and $P_\eta$ are Markov processes,

$$\begin{aligned}
\mathcal{F}_K(\eta; \mathbb{Q}) &= \sum_{k=0}^{K-1} \mathbf{E}_{P_\eta} \left[ \ln P_\mathbb{Q}(\boldsymbol{X}^{(t_{k+1})}|\boldsymbol{X}^{(t_k)}) \right] \\
&\quad - \sum_{k=0}^{K-1} \mathbf{E}_{P_\eta} \left[ \ln P_\eta(\boldsymbol{X}^{(t_k)}, \boldsymbol{X}^{(t_{k+1})}) \right] \\
&\quad + \sum_{k=1}^{K-1} \mathbf{E}_{P_\eta} \left[ \ln P_\eta(\boldsymbol{X}^{(t_k)}) \right]
\end{aligned}$$

We now express these terms as functions of $\mu_{\boldsymbol{x}}(t), \gamma_{\boldsymbol{x},\boldsymbol{y}}(t)$ and $q_{\boldsymbol{x},\boldsymbol{y}}$. By definition, $P_\eta(\boldsymbol{X}^{(t_k)} = \boldsymbol{x}) = \mu_{\boldsymbol{x}}(t_k)$. Each of the expectations either depend on this term, or on the joint distribution $P_\eta(\boldsymbol{X}^{(t_{k-1})}, \boldsymbol{X}^{(t_k)})$. Using the continuity of $\gamma_{\boldsymbol{x},\boldsymbol{y}}(t)$ we write

$$P_\eta(\boldsymbol{X}^{(t_k)} = \boldsymbol{x}, \boldsymbol{X}^{(t_{k+1})} = \boldsymbol{y}) = \delta_{\boldsymbol{x},\boldsymbol{y}}\mu_{\boldsymbol{x}}(t_k) + \Delta_K \cdot \gamma_{\boldsymbol{x},\boldsymbol{y}}(t_k) + o(\Delta_K)$$

where $\Delta_K = T/K$. Similarly, we can also write

$$P_{\mathbb{Q}}(\boldsymbol{X}^{(t_{k+1})} = \boldsymbol{y} | \boldsymbol{X}^{(t_k)} = \boldsymbol{x}) = \delta_{\boldsymbol{x},\boldsymbol{y}} + \Delta_K \cdot q_{\boldsymbol{x},\boldsymbol{y}} + o(\Delta_K)$$

Finally, using properties of logarithms we have that

$$\ln(1 + \Delta_K \cdot z + o(\Delta_K)) = \Delta_K \cdot z + o(\Delta_K).$$

Using these relations, we can rewrite after tedious yet straightforward manipulations,

$$\mathcal{F}_K(\eta; \mathbb{Q}) = \mathcal{E}_K(\eta; \mathbb{Q}) + \mathcal{H}_K(\eta),$$

where

$$\mathcal{E}_K(\eta; \mathbb{Q}) = \sum_{k=0}^{K-1} \Delta_K e_K(t_k), \quad \mathcal{H}_K(\eta) = \sum_{k=0}^{K-1} \Delta_K h_K(t_k),$$

and

$$e_K(t) = \sum_{\boldsymbol{x}} \sum_{\boldsymbol{y} \neq \boldsymbol{x}} \gamma_{\boldsymbol{x},\boldsymbol{y}}(t)[1 + \ln \mu_{\boldsymbol{x}}(t) - \ln \gamma_{\boldsymbol{x},\boldsymbol{y}}(t)] + o(\Delta_K)$$

$$h_K(t) = \sum_{\boldsymbol{x}} \left[\mu_{\boldsymbol{x}}(t) q_{\boldsymbol{x}\boldsymbol{x}} + \sum_{\boldsymbol{y} \neq \boldsymbol{x}} \gamma_{\boldsymbol{x},\boldsymbol{y}}(t) \log q_{\boldsymbol{x},\boldsymbol{y}}\right] + o(\Delta_K)$$

Letting $K \to \infty$ we have that $\sum_k \Delta_k[f(t_k) + o(\Delta_K)] \to \int_0^T f(t) dt$, hence $E_K(\eta; \mathbb{Q})$ and $\mathcal{H}_K(\eta)$ converge to $E(\eta; \mathbb{Q})$ and $\mathcal{H}(\eta)$, respectively. ∎

## 4 Factored Approximation

The variational principle we discussed is based on a representation that is as complex as the original process—the number of functions $\gamma_{\boldsymbol{x},\boldsymbol{y}}(t)$ we consider is equal to the size of the original rate matrix $\mathbb{Q}$. To get a tractable inference procedure we make additional simplifying assumptions on the approximating distribution.

Given a $D$-component process we consider approximations that factor into products of independent processes. More precisely, we define $\mathcal{M}_e^i$ to be the continuous Markov-consistent density sets over the component $X_i$, that are consistent with the evidence on $X_i$ at times 0 and $T$. Given a collection of density sets $\eta^1, \ldots, \eta^D$ for the different components, the product density set $\eta = \eta^1 \times \cdots \times \eta^D$ is defined as

$$\mu_{\boldsymbol{x}}(t) = \prod_i \mu_{x_i}^i(t)$$

$$\gamma_{\boldsymbol{x},\boldsymbol{y}}(t) = \begin{cases} \gamma_{x_i,y_i}^i(t) \mu_{\boldsymbol{x}}^{\setminus i}(t) & \delta(\boldsymbol{x},\boldsymbol{y}) = \{i\} \\ \sum_i \gamma_{x_i,x_i}^i(t) \mu_{\boldsymbol{x}}^{\setminus i}(t) & \boldsymbol{x} = \boldsymbol{y} \\ 0 & \text{otherwise} \end{cases}$$

where $\mu_{\boldsymbol{x}}^{\setminus i}(t) = \prod_{j \neq i} \mu_{x_j}^j(t)$ is the joint distribution at time $t$ of all the components other than the $i$'th. (It is not hard to see that if $\eta^i \in \mathcal{M}_e^i$ for all $i$, then $\eta \in \mathcal{M}_e$.) We define the set $\mathcal{M}_e^F$ to contain all factored density sets. From now on we assume that $\eta = \eta^1 \times \cdots \times \eta^D \in \mathcal{M}_e^F$.

Assuming that $\mathbb{Q}$ is defined by a CTBN, and that $\eta$ is a factored density set, we can rewrite

$$\mathcal{E}(\eta; \mathbb{Q}) = \sum_i \int_0^T \sum_{x_i} \mu_{x_i}^i(t) \mathbf{E}_{\mu^{\setminus i}(t)} \left[q_{x_i,x_i | \boldsymbol{U}_i}\right] dt$$

$$+ \sum_i \int_0^T \sum_{x_i, y_i \neq x_i} \gamma_{x_i, y_i}^i(t) \mathbf{E}_{\mu^{\setminus i}(t)} \left[\ln q_{x_i, y_i | \boldsymbol{U}_i}\right] dt,$$

and

$$\mathcal{H}(\eta) = \sum_i \mathcal{H}(\eta^i).$$

This decomposition involves only local terms that either include the $i$'th component, or include the $i$'th component and its parents in the CTBN defining $\mathbb{Q}$. Note that terms such as $\mathbf{E}_{\mu^{\setminus i}(t)}\left[q_{x_i, x_i | \boldsymbol{U}_i}\right]$ involve only $\mu^j(t)$ for $j \in \mathbf{Pa}_i$.

To make the factored nature of the approximation explicit in the notation, we write henceforth,

$$\mathcal{F}(\eta; \mathbb{Q}) = \mathcal{F}^F(\eta^1, \ldots, \eta^D; \mathbb{Q}).$$

**Fixed Point Characterization** We can now pose the optimization problem we wish to solve:

Fixing $i$, and given $\eta^1, \ldots, \eta^{i-1}, \eta^{i+1}, \ldots, \eta^D$, in $\mathcal{M}_e^1, \ldots \mathcal{M}_e^{i-1}, \mathcal{M}_e^{i+1}, \ldots, \mathcal{M}_e^D$, respectively, find $\arg\max_{\eta^i \in \mathcal{M}_e^i} \mathcal{F}^F(\eta^1, \ldots, \eta^D; \mathbb{Q})$.

If for all $i$, we have a $\mu^i \in \mathcal{M}_e^i$, which is a solution to this optimization problem with respect to each component, then we have a (local) stationary point of the energy functional within $\mathcal{M}_e^F$.

To solve this optimization problem, we define a Lagrangian, which includes the constraints in the form of Def. 3.1. The Lagrangian is a functional of the functions $\mu_{x_i}^i(t)$ and $\gamma_{x_i, y_i}^i(t)$ and Lagrange multipliers (which are functions of $t$ as well). The stationary point of the functional satisfies the *Euler-Lagrange* equations, namely the functional derivatives of $\mathcal{L}$ vanish. Writing these equations in explicit form we get a fixed point characterization of the solution in term of the following set of ODEs:

$$\frac{d}{dt}\mu_{x_i}^i(t) = \sum_{y_i \neq x_i} \left(\gamma_{y_i, x_i}^i(t) - \gamma_{x_i, y_i}^i(t)\right)$$

$$\frac{d}{dt}\rho_{x_i}^i(t) = -\rho_{x_i}^i(t)(\bar{q}_{x_i, x_i}^i(t) + \psi_{x_i}^i(t)) \quad (3)$$

$$- \sum_{y_i \neq x_i} \rho_{y_i}^i(t) \tilde{q}_{x_i, y_i}^i(t)$$

where $\rho^i$ are the exponents of the Lagrange multipliers.

In addition we have the following algebraic constraint

$$\rho^i_{x_i}(t)\gamma^i_{x_i,y_i}(t) = \mu^i_{x_i}(t)\tilde{q}^i_{x_i,y_i}(t)\rho^i_{y_i}(t), \quad x_i \neq y_i. \quad (4)$$

In these equations we use the following shorthand notations for the average rates

$$\bar{q}^i_{x_i,y_i}(t) = \mathbf{E}_{\mu^{\setminus i}(t)}\left[q^{i|\mathbf{Pa}_i}_{x_i,y_i|\mathbf{U}_i}\right]$$

$$\bar{q}^i_{x_i,y_i|x_j}(t) = \mathbf{E}_{\mu^{\setminus i}(t)}\left[q^{i|\mathbf{Pa}_i}_{x_i,y_i|\mathbf{U}_i} \mid x_j\right],$$

Similarly, we have the following shorthand notations for the geometrically-averaged rates,

$$\tilde{q}^i_{x_i,y_i}(t) = \exp\left\{\mathbf{E}_{\mu^{\setminus i}(t)}\left[\ln q^{i|\mathbf{Pa}_i}_{x_i,y_i|\mathbf{U}_i}\right]\right\}$$

$$\tilde{q}^i_{x_i,y_i|x_j}(t) = \exp\left\{\mathbf{E}_{\mu^{\setminus i}(t)}\left[\ln q^{i|\mathbf{Pa}_i}_{x_i,y_i|\mathbf{U}_i} \mid x_j\right]\right\},$$

The last auxiliary term is

$$\psi^i_{x_i}(t) = \sum_{j\in Children_i}\sum_{x_j}\mu^j_{x_j}(t)\bar{q}^j_{x_j,x_j|x_i}(t)+$$
$$\sum_{j\in Children_i}\sum_{x_j\neq y_j}\gamma^j_{x_j,y_j}(t)\ln\tilde{q}^j_{x_j,y_j|x_i}(t).$$

The two differential equations (3) for $\mu^i_{x_i}(t)$ and $\rho^i_{x_i}(t)$ describe, respectively, the progression of $\mu^i_{x_i}$ forward, and the progression of $\rho^i_{x_i}$ backward. To uniquely solve these equations we need to set the boundary conditions. The boundary condition for $\mu^i_{x_i}$ is defined explicitly in $\mathcal{M}^F_e$ as

$$\mu^i_{x_i}(0) = \delta_{x_i,e_{i,0}} \quad (5)$$

The boundary condition at $T$ is slightly more involved. The constraints in $\mathcal{M}^F_e$ imply that $\mu^i_{x_i}(T) = \delta_{x_i,e_{i,T}}$. As stated by Lemma 3.4, we have that $\gamma^i_{e_{i,T},x_i}(T) = 0$ when $x_i \neq e_{i,T}$. Plugging these values into (4), and assuming that $\mathbb{Q}$ is irreducible we get that $\rho_{x_i}(T) = 0$ for all $x_i \neq e_{i,T}$. In addition, we notice that $\rho_{e_{i,T}}(T) \neq 0$, for otherwise the whole system of equations for $\rho$ will collapse to 0. Finally, notice that the solution of (3) for $\mu^i$ and $\gamma^i$ is insensitive to the multiplication of $\rho^i$ by a constant. Thus, we can arbitrarily set $\rho_{e_{i,T}}(T) = 1$, and get the boundary condition

$$\rho^i_{x_i}(T) = \delta_{x_i,e_{i,T}}. \quad (6)$$

**Theorem 4.1:** $\eta^i \in \mathcal{M}^i_e$ is a stationary point (e.g., local maxima) of $\mathcal{F}^F(\eta^1,\ldots,\eta^D;\mathbb{Q})$ subject to the constraints of Def. 3.1 if and only if it satisfies (3–6).

It is straightforward to extend this result to show that at a maximum with respect to all the component densities, this fixed-point characterization must hold for all components simultaneously.

**Example 4.2:** Consider the case of a single component, for which our procedure should be exact, as no simplifying assumptions are made on the density set. In this case, the averaged rates $\bar{q}^i$ and the geometrically-averaged rates $\tilde{q}^i$ both reduce to the unaveraged rates $q$, and $\psi \equiv 0$. Thus, the system of equations to be solved is

$$\frac{d}{dt}\mu_x(t) = \sum_{y\neq x}(\gamma_{y,x}(t) - \gamma_{x,y}(t))$$

$$\frac{d}{dt}\rho_x(t) = -\sum_y q_{x,y}\rho_y(t),$$

along with the algebraic equation

$$\rho_x(t)\gamma_{x,y}(t) = q_{x,y}\mu_x(t)\rho_y(t), \qquad y \neq x.$$

In this case, it is straightforward to show that the backward propagation rule for $\rho_x$ implies that

$$\rho_x(t) = \Pr(e_T|X^{(t)}).$$

This system of ODEs is similar to forward-backward propagation, except that unlike classical forward propagation (which would use a function such as $\alpha_x(t) = \Pr(X^{(t)} = x|e_0)$), here the forward propagation already takes into account the backward messages, to directly compute the posterior. Given this interpretation, it is clear that integrating $\rho_x(t)$ from $T$ to $0$ followed by integrating $\mu_x(t)$ from $0$ to $T$ computes the exact posterior of the processes.

This interpretation of $\rho_x(t)$ also allows us to understand the role of $\gamma_{x,y}(t)$. Recall that $\gamma_{x,y}(t)/\mu_x(t)$ is the instantaneous rate of transition from $x$ to $y$ at time $t$. Thus,

$$\frac{\gamma_{x,y}(t)}{\mu_x(t)} = q_{x,y}\frac{\rho_y(t)}{\rho_x(t)}.$$

That is, the instantaneous rate combines the original rate with the relative likelihood of the evidence at $T$ given $y$ and $x$. If $y$ is much more likely to lead to the final state, then the rates are biased toward $y$. Conversely, if $y$ is unlikely to lead to the evidence the rate of transitions to it are lower. This observation also explains why the forward propagation of $\mu_x$ will reach the observed $\mu_x(T)$ even though we did not impose it explicitly.

**Example 4.3:** We define an *Ising chain* to be a CTBN $X_1 \leftrightarrow X_2 \leftrightarrow \cdots \leftrightarrow X_D$ such that each binary component prefers to be in the same state as its neighbor. These models are governed by two parameters: a *coupling parameter* $\beta$ which determines the strength of the coupling between two neighboring components, and a *rate parameter* $\tau$ that determines the propensity of each component to change its state. More formally, we define the conditional rate matrices as $q^{i|\mathbf{Pa}_i}_{x_i,y_i|\mathbf{u}_i} = \tau\left(1 + e^{-2y_i\beta\sum_{J\in\mathbf{Pa}_i}x_j}\right)^{-1}$ where $x_j \in \{-1,1\}$.

As an example, we consider a two-component Ising chain with initial state $X_1^{(0)} = -1$ and $X_2^{(0)} = 1$, and a reversed state at the final time, $X_1^{(T)} = 1$ and $X_2^{(T)} = -1$. For a large value of $\beta$, this evidence is unlikely as at both end points the components are in a undesired configurations. The exact posterior is one that assigns higher probabilities to trajectories where one of the components switches relatively fast to match the other, and then toward the end of the interval, they separate to match the evidence. Since the model is symmetric, these trajectories are either ones in which both components are most of the time in state $-1$, or ones where both are most of the time in state 1 (Fig. 1 top). Due to symmetry, the marginal probability of each component is around $0.5$ throughout most of the interval (Fig. 1 middle). The variational approximation cannot capture the dependency between the two components, and thus converges to one of two local maxima, corresponding to the two potential subsets of trajectories. Examining the value of $\rho^i$, we see that close to the end of the interval they bias the instantaneous rates significantly (Fig. 1 bottom).

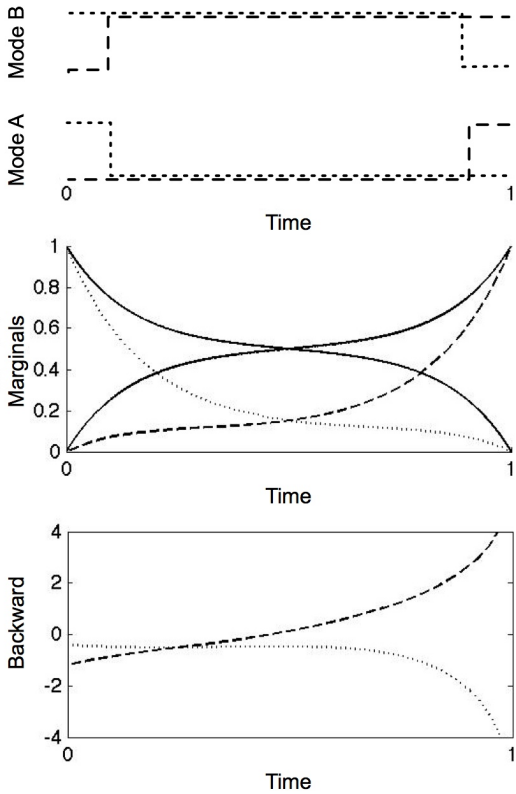

Figure 1: Numerical results for the two-component Ising chain described in Example 4.3 where the first component starts in state $-1$ and ends at time $T = 1$ in state 1. The second component has the opposite behavior. **(top)** Two likely trajectories depicting the two modes of the model. **(middle)** Exact (solid) and approximate (dashed/dotted) marginals $\mu_1^i(t)$. **(bottom)** The log ratio $\log \rho_1^i(t)/\rho_0^i(t)$.

This example also allows to examine the implications of modeling the posterior by inhomogeneous Markov processes. In principle, we might have used as an approximation Markov processes with homogeneous rates, and conditioned on the evidence. To examine whether our approx-

imation behaves in this manner, we notice that in the single component case we have

$$q_{x,y} = \frac{\rho_x(t)\gamma_{x,y}(t)}{\rho_y(t)\mu_x(t)},$$

which should be constant. Consider the analogous quantity in the multi-component case: $\tilde{q}^i_{x_i,y_i}(t)$, the geometric average of the rate of $X_i$, given the probability of parents state. Not surprisingly, this is exactly a mean field approximation, where the influence of interacting components is approximated by their average influence. Since the distribution of the parents (in the two-component system, the other component) changes in time, these rates change continuously, especially near the end of the time interval. This suggests that a piecewise homogeneous approximation cannot capture the dynamics without a loss in accuracy.

**Optimization Procedure** If $\mathbb{Q}$ is irreducible, then $\rho^i_{x_i}$ and $\mu_{x_i}$ are non-zero throughout the open interval $(0, T)$. As a result, we can solve (4) to express $\gamma^i_{x_i,y_i}$ as a function of $\mu^i$ and $\rho^i$, thus eliminating it from (3) to get evolution equations solely in terms of $\mu^i$ and $\rho^i$. Abstracting the details, we obtain a set of ODEs of the form

$$\frac{d}{dt}\mu^i(t) = \alpha(\mu^i(t), \rho^i(t), \mu^{\backslash i}(t)) \quad \mu^i(0) = \text{given}$$
$$\frac{d}{dt}\rho^i(t) = -\beta(\rho^i(t), \mu^{\backslash i}(t)) \qquad \rho^i(T) = \text{given}.$$

where $\alpha$ and $\beta$ can be inferred from (3) and (4). Since the evolution of $\rho^i$ does not depend on $\mu^i$, we can integrate backward from time $T$ to solve for $\rho^i$. Then, integrating forward from time 0, we compute $\mu^i$. After performing a single iteration of backward-forward integration, we obtain a solution that satisfies the fixed-point equation (3) for the $i$'th component. (This is not surprising once we have identified our procedure to be a variation of a standard forward-backward algorithm for a single component.) Such a solution will be a local maximum of the functional w.r.t. to $\eta^i$ (reaching a local minimum or a saddle point requires very specific initialization points).

This suggests that we can use the standard procedure of asynchronous update, where we update each component in a round-robin fashion. Since each of these single-component updates converges in one backward-forward step, and since it reaches a local maximum, each step improves the value of the free energy over the previous one. Since the free energy functional is bounded by the probability of the evidence, this procedure will always converge.

Another issue is the initialization of this procedure. Since the iteration on the $i$'th component depends on $\mu^{\backslash i}$, we need to initialize $\mu$ by some legal assignment. To do so, we create a fictional rate matrix $\tilde{\mathbb{Q}}_i$ for each component and initialize $\mu^i$ to be the posterior of the process given the evidence $e_{i,0}$ and $e_{i,T}$. As a reasonable initial guess, we choose at random one of the conditional rates in $\mathbb{Q}$ to determine the fictional rate matrix.

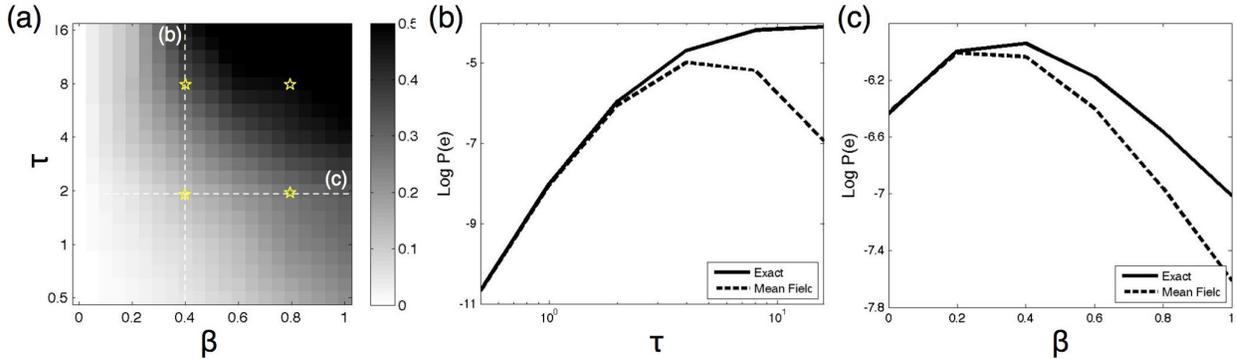

Figure 2: **(a)** Relative error as a function of the coupling parameter $\beta$ (x-axis) and transition rates $\tau$ (y-axis) for an 8-component Ising chain. **(b)** Comparison of true vs. estimated likelihood as a function of the rate parameter $\tau$. **(c)** Comparison of true vs. likelihood as a function of the coupling parameter $\beta$.

The continuous time update equations allow us to use standard ODE methods with an adaptive step size (here we use the Runge-Kutta-Fehlberg (4,5) method). At the price of some overhead, these procedure automatically tune the trade-off between error and time granularity.

## 5　Perspective & Related Works

Variational approximations for different types of continuous-time processes have been recently proposed [12, 13]. Our approach is motivated by results of Opper and Sanguinetti [12] who developed a variational principle for a related model. Their model, which they call a Markov jump process, is similar to an HMM, in which the hidden chain is a continuous-time Markov process and there are (noisy) observations at discrete points along the process. They describe a variational principle and discuss the form of the functional when the approximation is a product of independent processes. There are two main differences between the setting of Opper and Sanguinetti and ours. First, we show how to exploit the structure of the target CTBN to reduce the complexity of the approximation. These simplifications imply that the update of the $i$'th process depends only on its Markov blanket in the CTBN, allowing us to develop efficient approximations for large models. Second, and more importantly, the structure of the evidence in our setting is quite different, as we assume deterministic evidence at the end of intervals. This setting typically leads to a posterior Markov process in which the instantaneous rates used by Opper and Sanguinetti diverge toward the end point—the rates of transition into the observed state go to infinity, leading to numerical problems at the end points. We circumvent this problem by using the marginal density representation which is much more stable numerically.

Taking the general perspective of Wainwright and Jordan [15], the representation of the distribution uses the natural sufficient statistics. In the case of a continuous-time Markov process, the sufficient statistics are $T_{\boldsymbol{x}}$, the time spent in state $\boldsymbol{x}$, and $M_{\boldsymbol{x},\boldsymbol{y}}$, the number of transitions from state $\boldsymbol{x}$ to $\boldsymbol{y}$. In a discrete-time model, we can capture the statistics for every random variable. In a continuous-time model, however, we need to consider the time derivative of the statistics. Indeed, it is not hard to show that

$$\frac{d}{dt}\mathbf{E}\left[T_{\boldsymbol{x}}(t)\right] = \mu_{\boldsymbol{x}}(t) \quad \text{and} \quad \frac{d}{dt}\mathbf{E}\left[M_{\boldsymbol{x},\boldsymbol{y}}(t)\right] = \gamma_{\boldsymbol{x},\boldsymbol{y}}(t).$$

Thus, our marginal density sets $\eta$ provide what we consider a natural formulation for variational approaches to continuous-time Markov processes.

Our presentation focused on evidence at two ends of an interval. Our formulation easily extends to deal with more elaborate types of evidence: (1) If we do not observe the initial state of the $i$'th component, we can set $\mu_x^i(0)$ to be the prior probability of $X^{(0)} = x$. Similarly, if we do not observe $X_i$ at time $T$, we set $\rho_x^i(T) = 1$ as initial data for the backward step. (2) In a CTBN where one (or more) components are fully observed, we simply set $\mu^i$ for these components to be a distribution that assigns all the probability mass to the observed trajectory. Similarly, if we observe different components at different times, we may update each component on a different time interval. Consequently, maintaining for each component a marginal distribution $\mu^i$ throughout the interval of interest, we can update the other ones using their evidence patterns.

## 6　Experimental Evaluation

To gain better insight into the quality of our procedure, we performed numerical tests on models that challenge the approximation. Specifically, we use Ising chains where we explore regimes defined by the degree of coupling between the components (the parameter $\beta$) and the rate of transitions (the parameter $\tau$). We evaluate the error in two ways. The first is by the difference between the true log-likelihood and our estimate. The second is by the average relative error in the estimate of different expected sufficient statistics defined by $\sum_j \frac{|\hat{\theta}_j - \theta_j|}{\theta_j}$ where $\theta_j$ is exact value of the $j$'th ex-pected sufficient statistics and $\hat{\theta}_j$ is the approximation.

Applying our procedure on an Ising chain with 8 components, for which we can still perform exact inference, we evaluated the relative error for different choices of $\beta$ and $\tau$. The evidence in this experiment is $e_0 = \{+,+,+,+,+,+,-,-\}$, $T = 0.64$ and $e_T = \{-,-,-,+,+,+,+,+\}$. As shown in Fig. 2a, the error is larger when $\tau$ and $\beta$ are large. In the case of a weak coupling (small $\beta$), the posterior is almost independent, and our approximation is accurate. In models with few transitions (small $\tau$), most of the mass of the posterior is concentrated on a few canonical "types" of trajectories that can be captured by the approximation (as in Example 4.3). At high transition rates, the components tend to transition often, and in a coordinated manner, which leads to a posterior that is hard to approximate by a product distribution. Moreover, the resulting free energy landscape is rough with many local maxima. Examining the error in likelihood estimates (Fig. 2b,c) we see a similar trend.

Next, we examine the run time of our approximation when using fairly standard ODE solver with few optimizations and tunings. The run time is dominated by the time needed to perform the backward-forward integration when updating a single component, and by the number of such updates until convergence. Examining the run time for different choices of $\beta$ and $\tau$ (Fig. 3), we see that the run time of our procedure scales linearly with the number of components in the chain. Moreover, the run time is generally insensitive to the difficulty of the problem in terms of $\beta$. It does depend to some extent on the rate $\tau$, suggesting that processes with more transitions require more iterations to converge. Indeed, the number of iterations required to achieve convergence in the largest chains under consideration are mildly affected by parameter choices. The scalability of the run time stands in contrast to the Gibbs sampling procedure [4], which scales roughly with the number in transitions in the sampled trajectories. Comparing our method to the Gibbs sampling procedure we see (Fig. 4) that the faster Mean Field approach dominates the Gibbs procedure over short run times. However, as opposed to Mean Field, the Gibbs procedure is asymptotically unbiased, and with longer run times it ultimately prevails. This evaluation also shows that adaptive integration procedure in our methods strikes a better trade-off than using a fixed time granularity integration.

## 7  Inference on Trees

The abovementioned experimental results indicate that our approximation is accurate when reasoning about weakly-coupled components, or about time intervals involving few transitions (low transition rates). Unfortunately, in many domains we face strongly-coupled components. For example, we are interested in modeling the evolution of biological sequences (DNA, RNA, and proteins). In such systems, we have a *phylogenetic tree* that represents the branching process that leads to current day sequences (see Fig. 5a). It is common in sequence evolution to model this process as a continuous-time Markov process over a tree [6]. More precisely, the evolution along each branch is a standard continuous-time Markov process, and branching is modeled by a replication, after which each replica evolves independently along its sub-branch. Common applications are forced to assume that each character in the sequence evolves independently of the other.

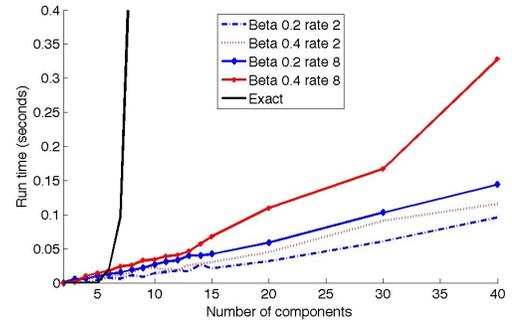

Figure 3: Evaluation of the run time of the approximation versus the run time of exact inference as a function of the number of components.

In some situations, assuming an independent evolution of each character is highly unreasonable. Consider the evolution of an RNA sequence that folds onto itself to form a functional structure. This folding is mediated by complementary base-pairing (A-U, C-G, etc) that stabilizes the structure. During evolution, we expect to see compensatory mutations. That is, if a $A$ changes into $C$ then its based-paired $U$ will change into a $G$ soon thereafter. To capture such coordinated changes, we need to consider the joint evolution of the different characters. In the case of RNA structure, the stability of the structure is determined by *stacking potentials* that measure the stability of two adjacent pairs of interacting nucleotides. Thus, if we consider a factor network to represent the energy of a fold, it will have structure as shown in Fig. 5b. We can convert this factor graph into a CTBN using procedures that consider the energy function as a fitness criteria in evolution [3, 16]. Unfortunately, inference in such models suffers from computational blowup, and so the few studies that deal with it explicitly resort to sampling procedures [16].

To consider trees, we need to extend our framework to deal with branching processes. In a linear-time model, we view the process as a map from $[0, T]$ into random variables $\boldsymbol{X}^{(t)}$. In the case of a tree, we view the process as a map from a point $\mathtt{t} = \langle \mathtt{b}, t \rangle$ on a tree $\mathcal{T}$ (defined by branch $\mathtt{b}$ and the time $t$ within it) into a random variable $\boldsymbol{X}^{(\mathtt{t})}$. Similarly, we generalize the definition of the Markov-consistent density set $\eta$ to include functions on trees. We define continuity of functions on trees in the obvious manner.

The variational approximation on trees is thus similar to the one on intervals. Within each branch, we deal with the same update formulas as in linear time. We denote by $\mu^i_{x_i}(\mathbf{b}, t)$ and $\rho^i_{x_i}(\mathbf{b}, t)$ the messages computed on branch b at time $t$. The only changes occur at vertices. Suppose we have a branch $\mathbf{b}_1$ of length $T_1$ incoming into vertex $v$, and two outgoing branches $\mathbf{b}_2$ and $\mathbf{b}_3$ (see Fig. 5c). Then we use the following updates for $\mu^i_{x_i}$ and $\rho^i_{x_i}$

$$\mu^i_{x_i}(\mathbf{b}_k, 0) = \mu^i_{x_i}(\mathbf{b}_1, T_1) \quad k = 2, 3,$$
$$\rho^i_{x_i}(\mathbf{b}_1, T_1) = \rho^i_{x_i}(\mathbf{b}_2, 0)\rho^i_{x_i}(\mathbf{b}_3, 0).$$

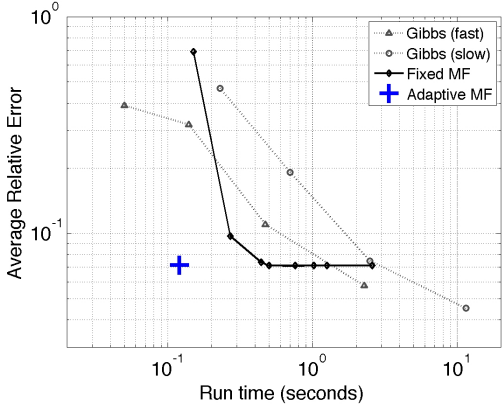

Figure 4: Evaluation of the run time vs. accuracy trade-off for several choices of parameters for Mean Field and Gibbs sampling on the branching process of Fig. 5(a).

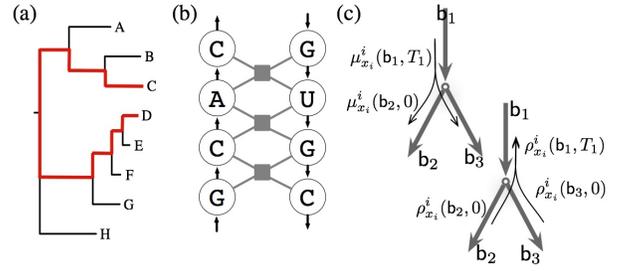

Figure 5: **(a)** An example of a phylogenetic tree. Branch lengths denote time intervals between events with the interval used for the comparison in Fig. 6a highlighted. **(b)** The form of the energy function for encoding RNA folding, superimposed on a fragment of a folded structure; each gray box denotes a term that involves four nucleotides. **(c)** Illustration of the ODE updates on a directed tree.

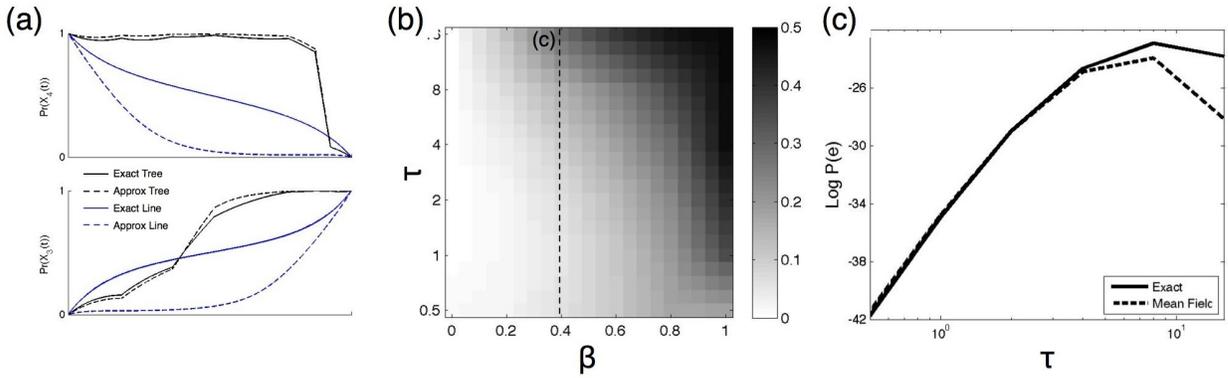

Figure 6: (a) Comparison of exact vs. approximate inference along the branch from $C$ to $D$ in the tree of Fig. 5(a) with and without additional evidence at other leaves. Exact marginals are shown in solid lines, whereas approximate marginal are in dashed lines. The two panels show two different components. (b) Evaluation of the relative error in expected sufficient statistics for an Ising chain in branching-time; compare to Fig. 2(a). (c) Evaluation of the estimated likelihood on a tree; compare to Fig. 2(b).

The forward propagation of $\mu^i$ simply uses the value at the end of the incoming branch as initial value for the outgoing branches. In backward propagation of $\rho^i$ the value at the end of $\mathbf{b}_1$ is the product of the values at the start of the two outgoing branches. This is the natural operation when we recall the interpretation of $\rho^i$ as the probability of the downstream evidence given the current state.

When switching to trees, we increase the amount of evidence about intermediate states. Consider for example the tree of Fig. 5a. We can view the span from $C$ to $D$ as an interval with evidence at its end. When we add evidence at the tip of other branches we gain more information about intermediate points between $C$ and $D$. To evaluate the impact of these changes on our approximation, we considered the tree of Fig. 5a, and compared it to inference in the backbone between $C$ and $D$ (Fig. 2). Comparing the true marginal to the approximate one along the main backbone (see Fig. 6a) we see a major difference in the quality of the approximation. The evidence in the tree leads to a much tighter approximation of the marginal distribution. A more systematic comparison (Fig. 6b,c) demonstrates that the additional evidence reduces the magnitude of the error throughout the parameter space.

As a more demanding test, we applied our inference procedure to the model introduced by Yu and Thorne [16] for a stem of 18 interacting RNA nucleotides in 8 species in the phylogeny of Fig. 5a. We compared our estimate of the expected sufficient statistics of this model to these obtained by the Gibbs sampling procedure. The results, shown in Fig. 7, demonstrate that over all, the two approximate inference procedures are in good agreement about the value of the expected sufficient statistics.

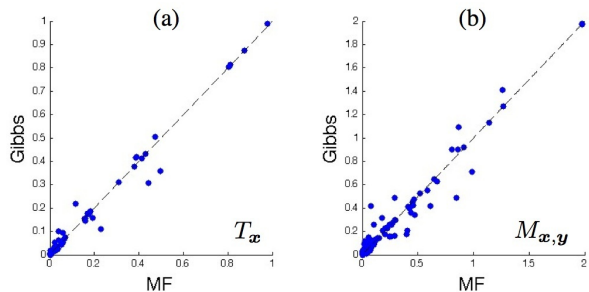

Figure 7: Comparison of estimates of expected sufficient statistics in the evolution of 18 interacting nucleotides, using a realistic model of RNA evolution. Each point is an expected statistic value; the $x$-axis is the estimate by the variational procedure, whereas the $y$-axis is the estimate by Gibbs sampling.

## 8 Discussion

In this paper we formulate a general variational principle for continuous-time Markov processes (by reformulating and extending the one proposed by Opper and Sanguinetti [12]), and use it to derive an efficient procedure for inference in CTBNs. In this mean field-type approximation, we use a product of independent inhomogeneous processes to approximate the multi-component posterior. Our procedure enjoys the same benefits encountered in discrete time mean field procedure [8]: it provides a lower-bound on the likelihood of the evidence and its run time scales linearly with the number of components. Using asynchronous updates it is guaranteed to converge, and the approximation represents a consistent joint distribution. It also suffers from expected shortcomings: there are multiple local maxima, and it cannot captures certain complex interactions in the posterior. By using a time-inhomogeneous representation, our approximation does capture complex patterns in the temporal progression of the marginal distribution of each component. Importantly, the continuous time parametrization enables straightforward implementation using standard ODE integration packages that automatically tune the trade-off between time granularity and approximation quality. We show how to extend it to perform inference on phylogenetic trees, and show that it provides fairly accurate answers in the context of a real application.

One of the key developments here is the shift from (piecewise) homogeneous parametric representations to continuously inhomogeneous representations based on marginal density sets. This shift increases the flexibility of the approximation and, somewhat surprisingly, also significantly simplifies the resulting formulation.

A possible extension of the ideas set here is to use our variational procedure to generate initial distribution for Gibbs sampling skip the initial burn-in phase and produce accurate samples. Another attractive aspect of this new variational approximation is its potential use for learning model parameters from data. It can be easily combined with the EM procedure for CTBNs [10], to obtain a Variational-EM procedure for CTBNs, which monotonically increases the likelihood by alternating between steps that improve the approximation $\eta$ (the updates discussed here) and steps that improve the model parameters $\theta$.

### Acknowledgments

We thank the anonymous reviewers for helpful remarks on previous versions of the manuscript. This research was supported in part by a grant from the Israel Science Foundation. Tal El-Hay is supported by the Eshkol fellowship from the Israeli Ministry of Science.

### References

[1] X. Boyen and D. Koller. Tractable inference for complex stochastic processes. In *UAI*, 1998.

[2] K.L. Chung. *Markov chains with stationary transition probabilities*. 1960.

[3] T. El-Hay, N. Friedman, D. Koller, and R. Kupferman. Continuous time markov networks. In *UAI*, 2006.

[4] T. El-Hay, N. Friedman, and R. Kupferman. Gibbs sampling in factorized continuous-time markov processes. In *UAI*, 2008.

[5] Y. Fan and C.R. Shelton. Sampling for approximate inference in continuous time Bayesian networks. In *AI and Math*, 2008.

[6] J. Felsenstein. *Inferring Phylogenies*. 2004.

[7] C.W. Gardiner. *Handbook of stochastic methods*. 2004.

[8] M. I. Jordan, Z. Ghahramani, T. Jaakkola, and L. K. Saul. An introduction to variational approximations methods for graphical models. In *Learning in Graphical Models*. 1998.

[9] U. Nodelman, C.R. Shelton, and D. Koller. Continuous time Bayesian networks. In *UAI*, 2002.

[10] U. Nodelman, C.R. Shelton, and D. Koller. Expectation maximization and complex duration distributions for continuous time Bayesian networks. In *UAI*, 2005.

[11] U. Nodelman, C.R. Shelton, and D. Koller. Expectation propagation for continuous time Bayesian networks. In *UAI*, 2005.

[12] M. Opper and G. Sanguinetti. Variational inference for Markov jump processes. In *NIPS*, 2007.

[13] C. Archambeau, M. Opper, Y. Shen, D. Cornford and J. Shawe-Taylor. Variational inference for Diffusion Processes. In *NIPS*, 2007.

[14] S. Saria, U. Nodelman, and D. Koller. Reasoning at the right time granularity. In *UAI*, 2007.

[15] M. J. Wainwright and M. Jordan. Graphical models, exponential families, and variational inference. *Found. Trends Mach. Learn.*, 1:1–305, 2008.

[16] J. Yu and J. L Thorne. Dependence among sites in RNA evolution. *Mol. Biol. Evol.*, 23:1525–37, 2006.